\documentclass[final,twocolumn]{elsarticle}

\usepackage{amssymb}

\usepackage{graphics} 
\usepackage{epsfig} 
\usepackage{xcolor}
\usepackage{amssymb}
\usepackage{adjustbox}
\usepackage{booktabs} 
\usepackage{amsmath}
\usepackage[ruled,vlined]{algorithm2e}
\usepackage{booktabs}
\usepackage{multirow}
\usepackage{gensymb}
\usepackage{makecell}
\usepackage{siunitx}
\usepackage{hyperref}
\usepackage[normalem]{ulem}

\usepackage{enumitem}

\newcommand{\CP}[1]{\ignorespaces}
\newcommand{\ie}{\textit{i.e.}}
\newcommand{\eg}{\textit{e.g.}}

\usepackage[switch,columnwise]{lineno}

{
  \par\vspace{\baselineskip}\noindent%
  \textbf{Responses to Reviewers}\begin{itshape}%
  \par\vspace{\baselineskip}\noindent%
}%
{\end{itshape}}

\journal{Computers and Electronics in Agriculture}

\begin{document}

\begin{frontmatter}

\title{Multispectral Vineyard Segmentation: A Deep Learning Comparison Study}

\author[inst1]{T. Barros\corref{cor1}} \ead{tiagobarros@isr.uc.pt} 
\author[inst1]{P. Conde} 
\author[inst2]{G. Gonçalves}
\author[inst1]{C. Premebida}
\author[inst1]{M. Monteiro}
\author[inst3,inst4,inst5]{ C.S.S. Ferreira} 
\author[inst1]{U.J. Nunes} 

 \cortext[cor1]{Corresponding author}

\affiliation[inst1]{organization={University of Coimbra, Institute of Systems and Robotics, Department of Electrical and Computer Engineering},
            city={Coimbra},
            country={Portugal}}
            


\affiliation[inst2]{organization={University of Coimbra,  Institute for Systems Engineering and Computers at Coimbra, Department of Mathematics},
            city={Coimbra},
            country={Portugal}}

\affiliation[inst3]{organization={Research Centre for Natural Resources, Environment and Society, Polytechnic Institute of Coimbra, Coimbra Agrarian Technical School},
            city={Coimbra},
            country={Portugal}}

\affiliation[inst4]{organization={Bolin Centre for Climate Research, Department of Physical Geography, Stockholm University},
            city={Stockholm},
            country={Sweden}}

\affiliation[inst5]{organization={Navarino Environmental Observatory},
            city={Navarino Dunes Messinia},
            country={Greece}}
            
\begin{abstract}
Digital agriculture has evolved significantly over the last few years due to the technological developments in automation and computational intelligence applied to the agricultural sector, including vineyards which are a relevant crop in the Mediterranean region.  In this work, a study is presented of semantic segmentation for vine detection in real-world vineyards by exploring state-of-the-art deep segmentation networks and conventional unsupervised methods. Camera data have been collected on vineyards using an Unmanned Aerial System (UAS) equipped with a dual imaging sensor payload, namely a high-definition RGB camera and a five-band multispectral and thermal camera. Extensive experiments using deep-segmentation networks and unsupervised methods have been performed on multimodal datasets representing four distinct vineyards located in the central region of Portugal. The reported results indicate that SegNet, U-Net, and ModSegNet have equivalent overall performance in vine segmentation. The results also show that multimodality slightly improves the performance of vine segmentation, but the NIR spectrum alone generally is sufficient on most of the datasets. Furthermore, results suggest that high-definition RGB images produce equivalent or higher performance than any lower resolution multispectral band combination. Lastly, Deep Learning (DL) networks have higher overall performance than classical methods. The code and dataset are publicly available on \url{https://github.com/Cybonic/DL_vineyard_segmentation_study.git}
\end{abstract}




\end{frontmatter}

\section{INTRODUCTION}

Deep Learning (DL) has been increasingly gaining relevance in precision agriculture, namely in remote sensing tasks. Remote sensing technology such as satellite and UAVs allow non-invasive and time-effective inspection techniques,  which enable the automation of tasks such as disease detection \cite{KERKECH2020105446}, crop yield prediction \cite{VANKLOMPENBURG2020105709}, and other monitoring-related tasks \cite{9213900}. Conversely to satellites, which are limited by temporal and resolution constraints, UAV-based remote sensing offers a cost-effective data collection approach to generate the necessary geospatial products of smaller crops such as vineyards \cite{Deng2018}. 


In vineyards,the use of UAV-based imagery combined with DL approaches enables the automation of complex tasks, such as: the inference of the spatio-temporal variability or the mapping the structure of vineyards;  tasks of particular relevance for designing site-specific management strategies \cite{DeCastro2018a}. These strategies minimize unnecessary treatments \cite{campos2019development} and, on the other hand, maximize both yield and quality \cite{Padua2020}.  However, to integrate such technology as a reliable source of information in a decision-making process, vine plants have to be discriminated from the remaining vegetation to avoid measurement contamination. Otherwise,  the farmers may be misled, causing poor decisions that may compromise yield. 

The most recurrent approaches for avoiding such contaminations resort to computer vision methods to perform row detection. These methods identify segments (or clusters) in images that contain only vine plants, which are used a posteriori in other tasks such as vigor maps or disease detection to extract only information from the pixels that belong to vine-rows. 

This work goes beyond row detection. Conversely, to traditional approaches, which perform row detection, this work resorts to DL-based segmentation approaches to detect vine plants. Specifically, the main goal of this work is to study the applicability of consolidated DL segmentation networks in a specific agricultural task such as vine segmentation using aerial imagery. From a computer vision perspective, this problem is relatively simple, given that the task in hand is a binary segmentation problem, where the positive class represents vine plants, and the negative is everything else. However, the adverse environmental conditions and the various growth stages of the plants over time, combined with a limited amount of available data, make the problem challenging.

In this context, this work presents a comprehensive study that was conducted with the following objectives: the first objective is to assess which bands or band combinations of a state-of-the-art multispectral (MS) sensor are more appropriate for this task; the second objective is to assess the relationship between resolution and performance, comparing for this purpose a high-definition RGB (HD-RGB) camera with a comparatively lower resolution MS sensor; and the third objective is to assess the appropriateness of DL-based segmentation approaches compared with classical methods, outlining their advantages and disadvantages.

\begin{figure}[tb]
    \centering
    \includegraphics[width=1\columnwidth]{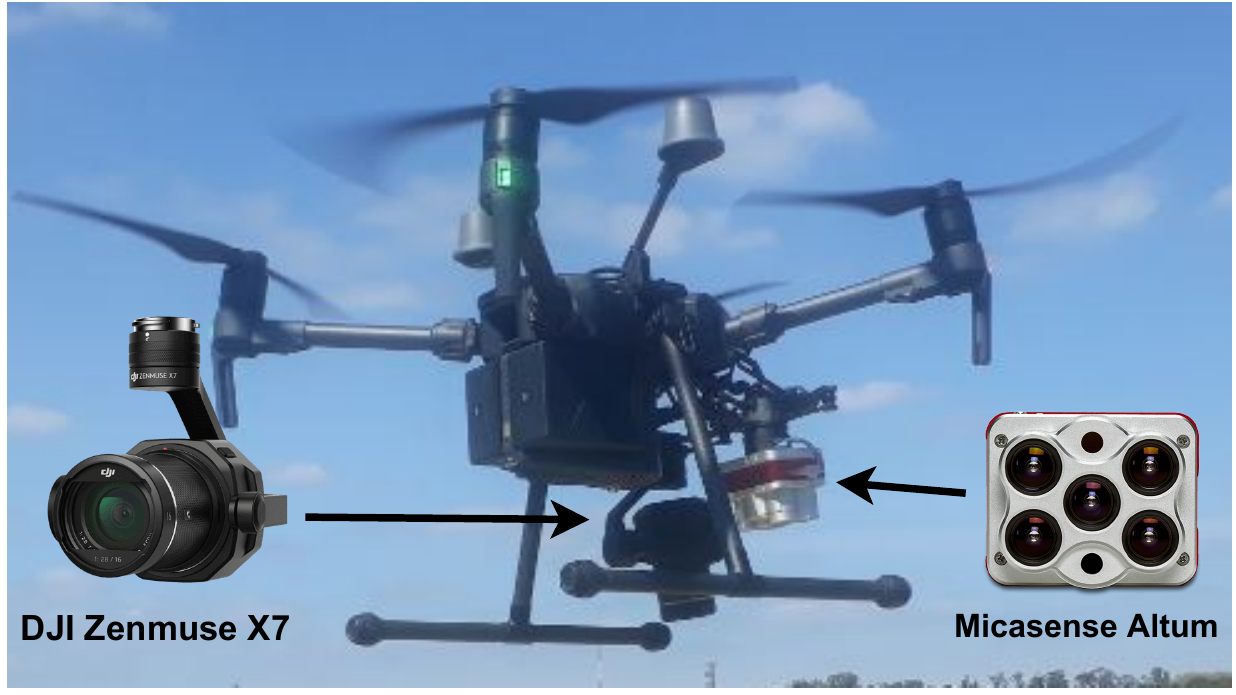}    
    \caption{UAS and the on-board cameras used for data collection. }
    \label{fig:UAS}
\end{figure}

To attain these objectives, this study was conducted on three state-of-the-art DL-based segmentation networks while using aerial imagery from three distinct vineyards captured by an UAV, with HD and MS sensors onboard (see Fig.\ref{fig:UAS}). All datasets used in this work are freely available, which we believe is a strong advantage for both DL and precision agriculture communities since there are very few aerial datasets available of vineyards comprising MS and HD-RGB orthomosaics, digital surface models, and ground-truth masks for segmentation.       

\par In summary, the main contributions of this work are the following:
\begin{itemize}[topsep=0pt]
     \setlength{\parskip}{0pt}
    \setlength{\itemsep}{0pt plus 1pt}
    \item A comparison study to assess the most appropriate spectral information for DL segmentation networks applied to the task of vine detection.
    \item[$\bullet$]  A new publicly available UAV-based vineyard dataset, with annotated segmentation masks, comprising MS,  HD-RGB orthomosaics, and digital surface models.
\end{itemize}

The remainder of this paper is organized as follows: Section\,\ref{sec:relatedwork} presents the state-of-the-art in the domain of semantic segmentation using UAV/drone data for precision agriculture, namely applied to vineyards. 

Section\,\ref{sec:mm} describes the material and methods used to conduct the study, detailing the framework, the methods and respective tools related to the dataset acquisition, and the techniques applied to perform segmentation. Section \ref{sec:experiments} highlights the implementation details of the experimental evaluation. Section \ref{sec:results} reports and discusses the results. Finally, section \ref{sec.conclusion} concludes the findings of this study and suggests future research directions.

\begin{figure*}[t]
    \centering
    \includegraphics[width=1\textwidth]{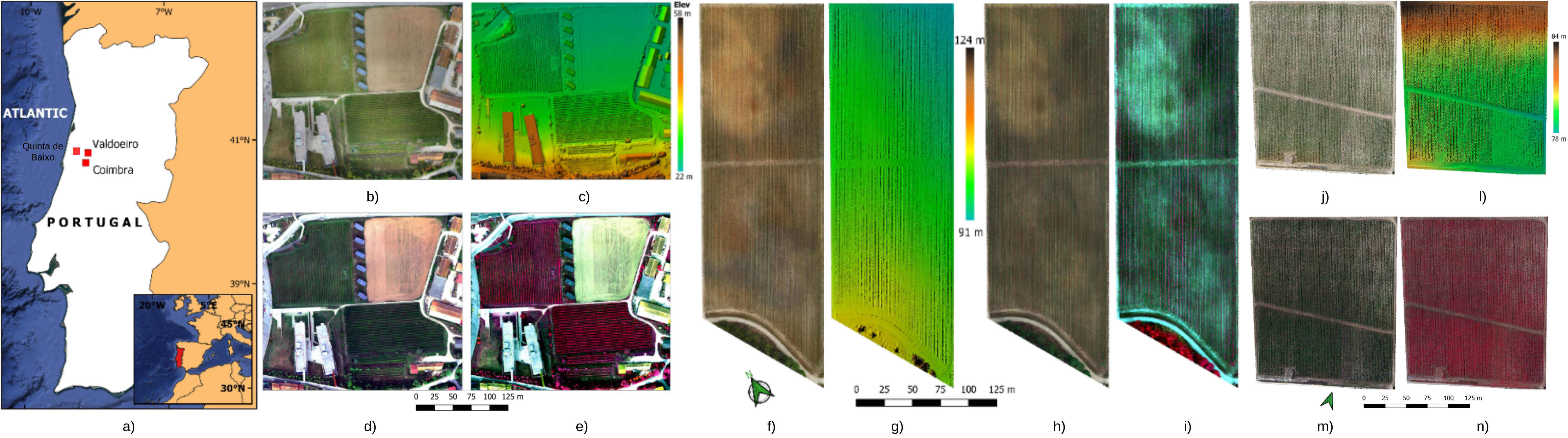}    
    \caption{Overview of the geographic locations of the three vineyard sites and their corresponding orthomosaics. The vineyard locations are shown in a). The orthomosaics of Esac are shown in the following sub-figures: b) corresponding to the orthomosaic based on HD images, captured by the X7 sensor; c) representing the digital surface model (DSM); d) corresponding to the R-G-B composition of the MS images captured by the Micasense Altum sensor; d) corresponding to the false-color RE-R-G composition also captured by the Micasense Altum sensor. The orthomosaics of Valdoeiro are shown in the following sub-figures: f) representing the HD-based orthomosaic; g) the DSM; h) the R-G-B composition; i)  the false-color RE-R-G composition. The orthomosaics of Quinta de Baixo are shown in the following sub-figures: j) corresponding to the HD-based orthomosaic; l) the DSM; m) the R-G-B composition; n) the false-color RE-R-G composition.
    }
    \label{fig:vineyard_overview}
\end{figure*}
\section{RELATED WORK}
\label{sec:relatedwork}

\begin{table*}[t]
  \centering
  
  \caption{Related work on  mulstispectral data for semantic segmentation in digital/precision agriculture.}
  \label{sota}
  \begin{adjustbox}{max width=\textwidth}
    \begin{tabular}{ccccp{0.35\textwidth}}
    \toprule
    \multicolumn{1}{c}{\textbf{Ref}} & \multicolumn{1}{c}{\textbf{Bands/Data Type}} & \multicolumn{1}{c}{\textbf{Fusion}} & \multicolumn{1}{c}{\textbf{Architecture/Approach}} & \multicolumn{1}{c}{\textbf{Application}} \\
    \midrule
    \midrule
     \cite{9213900} & RGB   & \thead{Late\\(HSV)}  & \thead{Otsu’s thresholding\\ Hough Transformation} & Vineyard row detection\\
    \cite{romero2018vineyard} & RGB+NIR+RE & \thead{Early\\(Vegetation Indices)} &   \thead{Two-layer\\feedforward network} & Vineyard water status estimation \\
    \cite{HALL2003813} &  NIR+RGB & \thead{Early\\(NDVI)} & Histogram & Vineyard canopy characterising and mapping  \\ 
    \cite{ahmed2019automatic} & RGB+NIR+RE & \thead{Early\\(NDVI)} & \thead{Laplacian of Gaussian\\unsupervised clustering \\ random walker} & Detection and segmentation of lentil plots \\
    \cite{COMBA201578} & RGB & \thead{Early\\(Gray-scale)} &
    \thead{Hough Space Clustering \\Total Least Squares} & Vineyard detection \\ 
    \cite{comba2018unsupervised} & Point Clouds  & -- & Unsupervised  &  Vineyard detection\\ \hline
    \cite{KERKECH2020105446} & RGB+NIR & \thead{Late\\(case-based)} & \thead{Encoder-Decoder\\ (SegNet)}  & Mildew disease detection in vine + row detection \\
    \cite{fawakherji2019crop} & \multicolumn{1}{c}{RGB} & \thead{Early\\(Concatenation)}  & \thead{Encoder-Decoder\\(Unet)} & Weed/crop segmentation and classification  \\
    \cite{bah2019crownet} & RGB   & \thead{Early\\(Concatenation)} &\thead{ S-SegNet\\ HoughCNet} & Crop row detection \\
    \cite{song2020identifying} & RGB+NIR & \thead{Early\\(Concatenation)} & \thead{Encoder-Decoder\\(SegNet)} & Identification of sunflower lodging \\
    \cite{kerkech2020vine} & (RGB+NIR)\newline{} + DSM & \thead{Late\\(case-based)} & \thead{Encoder-Decoder\\ (SegNet)} & Vine disease detection + row detection \\
    \bottomrule
    \end{tabular}%
  \end{adjustbox}
\end{table*}%

Precision agriculture, in general, has greatly benefited from the advances of machine learning and remote sensing, namely using multispectral (MS) sensors. These sensors can capture relevant information regarding biological phenomenons in plants that are not captured by the RGB spectrum.

In vineyards, MS information is widely used in many applications, using the data either from UAVs or satellites. In \cite{cogato2020medium}, the spectral bands of Sentinel-2 are used to assess the vineyards' damage and recovery time after a late frost event.  Despite the evident advantages of satellite-based sensing in agriculture, the specific case of vineyards is particularly challenging for many of these systems (including Sentinel) because of its low spatial resolution (10-50m/pixel) when compared with the 2\,m (approximately) of the inter-row distance in vineyards.  With such resolution, one pixel may represent a crop area that comprises multiple rows and thus making it difficult to discriminate between inter-row plants (\eg., weeds) and vine plants. Consequently, this leads to measurements contamination \cite{khaliq2019comparison}.

UAVs, on the other hand,  are more flexible and adjustable in altitude to obtain adequate image resolution.  Works have been using UAVs equipped with MS or/and RGB cameras to collect field data from crops. In vineyards, UAVs are frequently used to collect data for disease detection \cite{KERKECH2020105446}, water status assessment of vine plants as in \cite{romero2018vineyard}, among other applications. Recent research has shown that the primary information source in these domains is provided by RGB, Red-Edge (RE), and near-infrared (NIR) bands. Survey data can be used to generate geospatial products such as digital surface models (DSM), which are used as simple dept maps as an additional source of information \cite{KERKECH2020105446}.   It is interesting to note, in this context, that most of the UAVs are equipped only with one camera, either a MS sensor or an RGB camera while, in our work, we equipped our UAV with a dual gimbaled sensor system, combining a state-of-art HD-RGB camera and a MS sensor.
 
MS aerial imagery provides both rich spectral and spatial information. However, in the vineyard context, only the pixels that belong to the vine plant are of interest. Approaches to assess these pixels have differed over the years. A common one - still today - is to convert spectral information to vegetation indices (\eg, NDVI) and thereof resort to a semantic segmentation technique to identify pixels belonging to the vine plants. 
Early (classical) segmentation approaches were mainly based on thresholds \cite{9213900}, color indices \cite{kirk2009estimation}, clustering  \cite{COMBA201578}, histograms \cite{HALL2003813} or classical supervised  \cite{guerrero2012support} and unsupervised \cite{comba2018unsupervised} learning  methods.   
Advantages of these classical approaches include simplicity, `shallow' training, and low computation cost. On the other hand, the disadvantages, particularly in the agriculture context, are mainly related to low performance when faced with different lighting conditions, shadows, or complex backgrounds, making them more suitable for simpler and non-changing environments. A survey on early segmentation approaches in agriculture can be found in \cite{HAMUDA2016184}.     
 
More recent works resort to DL techniques, which have created a new momentum in many scientific areas, including digital agriculture, where many of the algorithms rely on Convolutional Neural Networks (CNNs) to learn features from the input representation.  In the agricultural domain, works related to segmentation rely on deep-networks, for example, the encoder-decoder SegNet \cite{badrinarayanan2017segnet} and U-Net \cite{ronneberger2015u}. In \cite{bah2019crownet}, authors propose CRowNet, which relies on S-SegNet \cite{badrinarayanan2017segnet} and a CNN-based Hough transform CNNs for row detection in RGB images. In  \cite{kerkech2020vine} and \cite{KERKECH2020105446},  SegNet is used for disease detection using as input RGB, NIR, and DSM, in the former, and RGB plus NIR in the latter.  In other crops such as sunflowers, the RGB and NIR bands are also used as inputs to a SegNet\cite{song2020identifying}. In \cite{fawakherji2019crop}, a U-Net is used for crop/weed classification. In our work, we make use of U-Net and SegNet, as well as an additional model called ModSegNet \cite{ganaye2018semi}, which is also an encoder-decoder network, to compare their performances when using imagery from a HD-RGB camera and a MS sensor on a vineyard segmentation task.


To summarise the related work on semantic segmentation applied to precision agriculture, particularly for vineyard-row detection, Table \ref{sota} presents a comprehensive view of the S.O.T.A. highlighting the classical $vs$ DL-based methods, the spectral bands and data representation, the fusion strategies, followed by the related architectures/models that have been used in the application domains.

\begin{figure*}[t]
    \centering
    \includegraphics[width=1\textwidth]{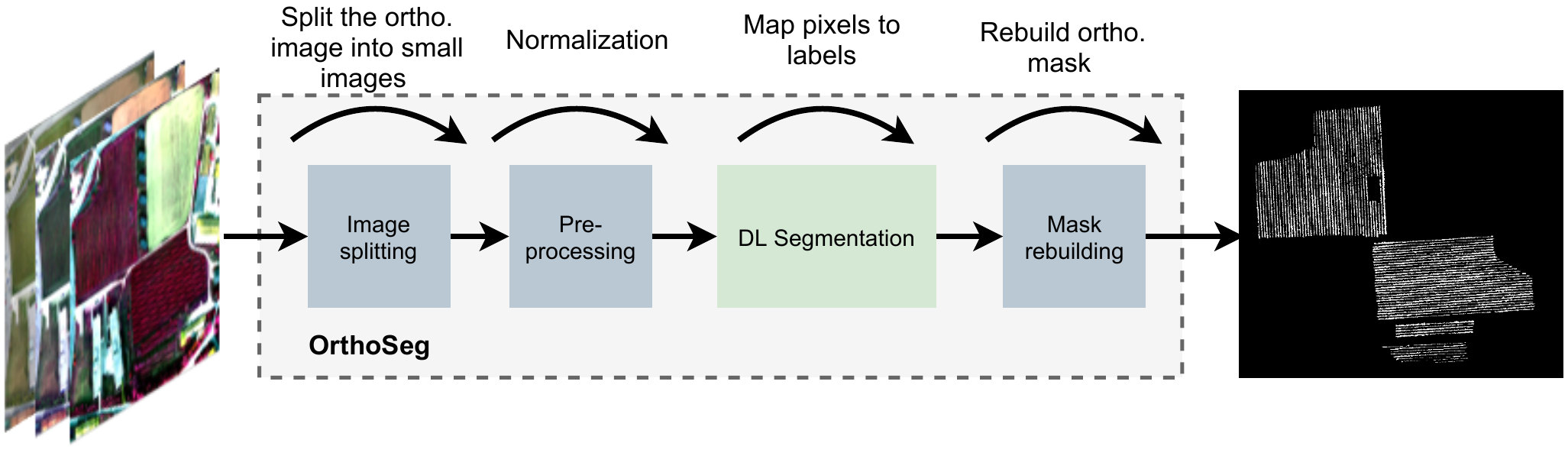}
    \caption{Orthomosaic segmentation pipeline (OthoSeg) with the following modules: image splitting, which splits the orthomosaics into sub-images; pre-processing, which normalizes each band of the sub-images; DL segmentation, which predicts sub-masks using a DL-based segmentation approach; and mask rebuilding, which uses the sub-masks to build a mask with the same size of the input orthomosaic.}
    \label{fig:pipeline}
\end{figure*}

\begin{figure}[htb]
    \centering
    \includegraphics[width=1\columnwidth,trim={0.5cm 0.1cm 0.2cm 0.3cm},clip]{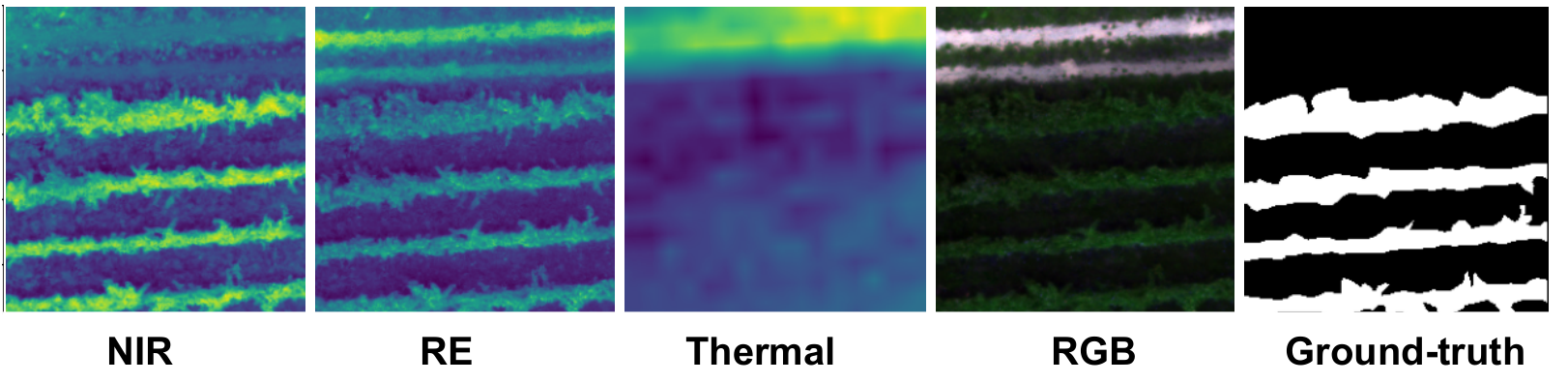}    
    \caption{Image examples of the Vineyards showing the spectral bands that integrate the mustispectral sensor, and a ground-truth mask.}
    \label{fig:ms_bands}
\end{figure}

\section{MATERIALS AND METHODS} \label{sec:mm}

In this section, the methods, framework, processes, and the `tools' that have been used in this study are described.  The first part is related to the field data, which includes the characterization of the study sites, as well as the description of the materials and data acquisition process. The second part of this section is dedicated to the methods used to obtain the results, namely the OrthoSeg pipeline (shown in Fig\,\ref{fig:pipeline}), where a comprehensive description of the various stages is presented, including the three DL models, and then we present a description of the classical segmentation methods that have been used for comparison purposes.

\subsection{Study Sites}

The study was carried out in three vineyards located in the Centre of mainland Portugal. Two vineyards, designated hereafter by Valdoeiro and Quinta de Baixo, are located in the Bairrada wine region, while the third vineyard (ESAC) is a ``living-lab/farm" within the Agrarian School of Coimbra (Fig.\,\ref{fig:vineyard_overview}). All the studied vineyards belong to a region with a Mediterranean climate, subjected to a strong influence of the Atlantic Ocean, characterized by average annual rainfall of 1077 mm and average annual temperature of \SI{15}{\celsius} \cite{ferreira2018runoff}, marked by a relatively long and dry summer (June-August). All vineyards are managed under conventional practices but present different biophysical characteristics.

Valdoeiro is a 2.9 ha vineyard, located at an altitude of 99 m, in flat terrain ($<2\degree$) under Cambisoil soil type, with a northeast-southwest exposure. The vineyard was planted in 2005 with a typical Baga vine variety and an approximate density of 3200 vines per ha, with plants spacing 1.3\,m in straight rows, an inter-rows distance of 2.4\,m, and a row azimuth of approximately 210\degree.

Quinta de Baixo covers an area of 3.2\,ha, located at an average altitude of 90\,m, in a smoothly sloping terrain (2\degree-5\degree), under Podzol soil type. The vineyard was planted in 2002, with Syrah, Pinot and Baga vine varieties with a density of 4400 vines per ha. The vines are installed at 0.9\,m apart within the rows, 1.1\,m between rows, and row azimuth of approximately 162\degree.

The ESAC vineyard extends over an area of 2.3\,ha divided into two plots: Esac1 and Esac2 (see Fig.\,\ref{fig:learning_data}.a). These plots are located at an altitude of 28\,m, in a smoothly sloping terrain (2\degree-5\degree) under Fluvisol soil type. 
The vineyard was planted in 1999 with different vine varieties such as Alfrocheiro, Aragonez, Touriga Nacional, and Marselan. Esac1 has a south-north exposure with an approximate plant density of 2800 vines per ha, a plant distance of 1.5\,m in straight rows, an inter-rows distance of 2.4\,m, and row azimuth of approximately 177\degree.  Esac2 has an east-west solar exposure with a plant density of approximately 3400 vines per ha, a plant distance of 1.4 m, an inter-row distance of 2.1\,m, and a row azimuth of approximately 266\degree.


\subsection{Materials and Data Acquisition} %

\begin{table*}[t]
\caption{Specifications of the two sensors integrated in the dual imaging payload. Field of view (FoV), Ground Sample Distance (GSD).}
\label{tab:sensors}
{\renewcommand{\arraystretch}{1.5}
\begin{adjustbox}{max width=\textwidth}
\begin{tabular}{cccccc}
\toprule
\textbf{Sensor} & \textbf{Band: Center wavelength (width) [nm]} &  \textbf{Resolution [px]} & \textbf{Focal Length [mm]} & 
\textbf{FoV [\degree]} & \textbf{\thead{GSD@100m \\AGL[cm/px]}} \\
\midrule
\midrule
\multirow{2}[1]{*}{Micasense Altum}& 
\thead{ B: 475 (32); G: 560 (27); R: 668 (16); RE: 717 (12); NIR: 840 (57)} &   
2064x1544 &  8 & 48 $\times$ 37  & 4\\
 &  
 Thermal: 11000 (60)  & 57 $\times$ 44 & 1.77 & 57x44 &  \multicolumn{1}{c}{67} \\\hline
DJI Zenmuse X7 (24) &  R,G,B & 6016 $\times$ 4008 (3:2)  &  24 &  52.2 $\times$ 36.2  & \multicolumn{1}{c}{1.6}
\\\bottomrule
\end{tabular}
\end{adjustbox}
}
\end{table*}

\begin{table*}[t]
  \centering
  \caption{UAS surveys and the corresponding GSD of the generated geospatial products.}
  \label{tab:geospatial}%
  {\renewcommand{\arraystretch}{1.5}
  \begin{adjustbox}{max width=\textwidth}
    \begin{tabular}{cccccc|ccc}
    \toprule
    \multirow{2}[2]{*}{\textbf{Location}} & \multicolumn{1}{c}{\multirow{2}[2]{*}{\textbf{Date \newline{} [mm/dd/yyyy]}}} & \multicolumn{1}{c}{\multirow{2}[2]{*}{\textbf{Time \newline{} [hh:mm]}}} & \multicolumn{1}{c}{\multirow{2}[2]{*}{\textbf{Duration \newline{}[min]}}} & \multirow{2}[2]{*}{\textbf{Weather}} & \multicolumn{1}{c|}{\multirow{2}[2]{*}{\textbf{ Flying height [m] AGL}}} & \multicolumn{3}{c}{\textbf{GSD [cm/pix]}} \\
          &       &       &       &       &       & \textbf{RGB } & \textbf{Multispectral} & \textbf{DSM} \\
    \midrule
    \midrule
    Coimbra & 10/01/2020 & 1:40:00 pm & 17    & Sunny & 120   & 1.7   & 4.8   & 3.4 \\
    \midrule
    Valdoeiro & 04/15/2021 & 11:45:00 am & 10    & Sun/cloud & 60    & 1     & 3     & 2 \\
    \midrule
     Quinta de Baixo &  07/27/2021 &  12:10:00 pm& 15   & Sunny & 60 & 1 &  2.5 &  2 \\
    \bottomrule
    \end{tabular}%
\end{adjustbox}
   }
\end{table*}%

To survey the study areas, a compact and ``low-cost" UAS from DJI (shown in Fig. \ref{fig:UAS}) was equipped with a MS sensor (Micasense Altum), a HD-RGB camera (Zenmuse X7), and a global navigation satellite systems (GNSS) with RTK correction. The UAS's flight missions have been planned with the DJI Pilot 1.9 software, where the front and side overlap was set to 80\% and 70\%, respectively, using the Altum sensor as a reference. 

The Altum sensor captures four spectral bands (R, G, B, RE, NIR) with a 2064 $\times$ 1544 resolution and a thermal band with a lower resolution of 57$\times$44;  a sample of each band is illustrated in Fig. \ref{fig:ms_bands}. The Zenmuse X7 sensor captures R, G, and B bands with a resolution of 6016 $\times$ 4008. A more detailed description of the cameras is provided in Table\,\ref{tab:sensors}.

The data acquisition process was carried out by surveying all sides (\ie, ESAC, Valdoeiro, and Quinta de Baixo) with custom settings, which were set to optimize information acquisition at survey time. Namely, the altitude at which the vineyard plots were surveyed was adjusted at each site. The Coimbra plots were surveyed in October at an altitude of 120\,m after the harvest was finished. The Valdoeiro plot was surveyed in April. At this time, the plants are still in an early growth stage with no, or few, visible leaves which makes plant recognition difficult at 120\,m. Thus, the height was adjusted to 60\,m to capture more rich and detailed information from the plants. The Quinta de Baixo vineyard was surveyed at 70\,m at the end of July, which is a critical season for vineyards, since plants are very advanced in the growth stage and diseases are more prevalent.

After data acquisition, raw images of both cameras were used to generate the geospatial products (\ie, DSM and orthomosaic) of the respective vineyards. 
The HD images were used to generate both the HD orthomosaics and the DSMs of the vineyards, which were computed based on a workflow presented in \cite{Goncalves2021}. The MS orthomosaics, on the other hand, were generated using the MS images acquired by the Altum sensor and computed based on the workflow proposed by Agisoft Metashape Professional Edition software (Agisoft LLC, St. Petersburg, Russia) version 1.7.2. However, before using such workflow, the MS images were pre-processed in order to apply the necessary radiometric corrections: vignetting, dark pixel offset, and converting raw images to radiance and then to reflectance space. 

The conversion process resorted to pre-and post-flight images from an Altum calibration panel, which was located in vineyards for adequate reflectance calibration. Furthermore, when the illumination conditions changed over time (due to sun/cloud conditions), an additional correction step has been performed using the Downwelling Light Sensor (DLS).  An overview of the survey conditions and the geospatial products are presented in Table \ref{tab:geospatial}, while the generated geospatial products are presented in Fig.\,\ref{fig:vineyard_overview}. 


\subsection{Orthomosaic Deep Learning-based Segmentation}

Orthomosaics are data structures that may have a large and arbitrary size. Such data structures are not appropriate to feed directly to DL-based approaches, which rely on CNN and are optimized for grid-based and fixed-sized inputs. Moreover, the computational demands of CNNs increase proportionally with the input size, which makes feeding orthomosaics directly to DL networks computationally too expensive.  To overcome this limitation, this work resorts to an approach (named OrthoSeg illustrated in Fig. \ref{fig:pipeline}) that has the following steps: receives orthos as inputs; splits these orthos into sub-images and then pre-processes the sub-images; the pre-processed sub-images are fed to the segmentation network, which outputs prediction sub-masks. Finally, the sub-masks are rebuilt into an orthomosaic mask of the same size as the input. 

\subsubsection{Orthomosaic Splitting \& Rebuilding}

The image splitting approach has been devised to divide the orthomosaics of all bands into smaller
sub-images with a fixed size of 240$\times$240 pixels, which represents a much less computational burden for DL segmentation networks.

\begin{figure}[t]
    \centering
    \includegraphics[width=1\columnwidth,trim={0.3cm 0.5cm 0.3cm 0.3cm},clip]{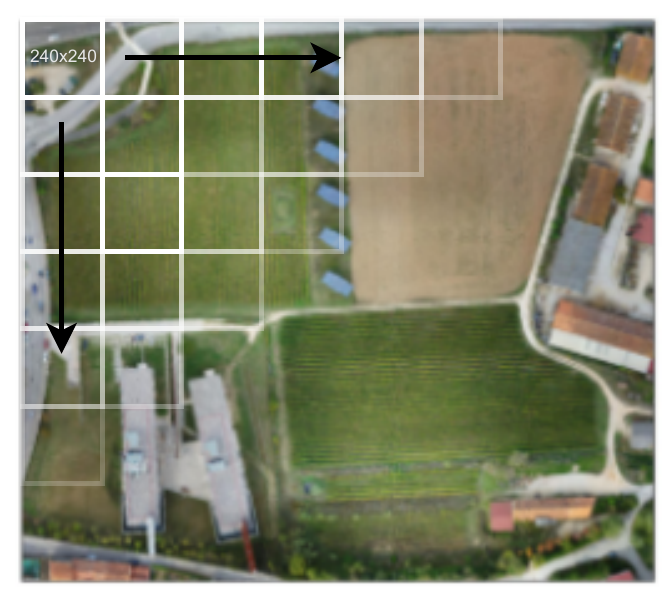}    
    \caption{Orthomosaic splitting approach. The splitting begins at the upper left corner and proceeds to the right until the end of the row. The process is repeated until the bottom.}
    \label{fig:split}
\end{figure}

The splitting process, illustrated in Fig. \ref{fig:split}, begins at the top-left corner of the orthomosaic and proceeds to the right, creating sub-images every 240 pixels. After the row is completed, a new row is defined 240 pixels below.  The process is repeated until the whole orthomosaic has been processed. 

\subsubsection{Pre-processing}

In order to improve convergence at training time, the generated sub-images are standardized using (\ref{eq:std}), before being fed to the neural network,
\begin{equation}
     X'_{b} = \frac{X_{b}-\mu_b}{\sigma_b}
     \label{eq:std}
\end{equation}

\noindent where $X_b$ represents the sub-image of the band $b$, $\mu_b$ is the mean, $\sigma_b$ denotes the standard deviation and $X'_b$ the corresponding standardized sub-image.

\subsubsection{Deep Segmentation Networks}

In this work, three deep neural networks are used for the task of supervised semantic segmentation:  U-Net \cite{ronneberger2015u},  SegNet \cite{badrinarayanan2017segnet} and ModSegnet \cite{ganaye2018semi}. All three networks are state-of-the-art DL-based segmentation approaches, particularly SegNet and U-Net that have been widely used in various fields, including agriculture. The three networks have a similar encoder-decoder-like architecture followed by a pixel-wise classification layer.  A compact formulation  of such networks can be expressed as follows:

\begin{equation} \label{eq:enc_dec}
    F_m =   \text{Dec}(\text{Enc}(I,\theta_{\text{Enc}}),\theta_{\text{Dec}})
\end{equation}
\begin{equation} \label{eq:classifier}
    M =  \text{C}(F_m)
\end{equation}

\noindent where Enc($\cdotp$) is the encoder, which receives as input parameters the encoder's weights $\theta_{\text{Enc}}$ and an image ($I \in \mathbb{R}^{b \times h \times w}$), where $b$, $h$ and $w$ represent respectively, the number of spectral bands, image height, and image width. The decoder Dec($\cdotp$) has as input parameters the encoder's outputs and the decoder's weights $\theta_{\text{Dec}}$. The Decoder outputs a feature map ($F_m \in \mathbb{R}^{l \times h \times w}$), with $l$ representing the number of classes, that is fed to a classifier C($\cdotp$). The classifier  outputs a mask ($M \in \mathbb{R}^{c \times h \times w}$) where $c$ represents the number of existing classes.

In the encoder, the reduction is performed by consecutive operators, where each includes a $3 \times 3$ unpadded convolution layer, a batch normalization (BatchNorm) layer \cite{ioffe2015batch}, a rectified linear unit (ReLu) layer and a dropout layer \cite{srivastava2014dropout}; each of these operators is followed by a $2 \times 2$ max-pooling layer to achieve translation invariance over small spatial shifts. The decoder uses the same consecutive sets of operations, in this case followed by an upsampling operation which transforms spatially $F_m$ to $M$.

\indent The SegNet architecture uses the same indices of the max pooling operations learned in the respective encoder steps, avoiding the need to learn new indices for the upsampling phase. On the other hand, U-net learns new indices for the transposed convolution operations used for the upsampling but has the particularity of concatenating each new feature space, obtained after each upsampling step, with the cropped feature space from the end of the corresponding encoder stage. ModSegnet incorporates both the memorized polling indices and the concatenation of feature spaces from the aforementioned architectures. Finally, in the last step of each architecture, we use a $1\times1$ convolution to map the final feature space to a prediction mask with the same size as the input image.

\subsection{Unsupervised Segmentation Methods}

Two unsupervised segmentation methods, K-means \cite{lloyd1982least} and OTSU \cite{otsu1979threshold}, are used for comparison purposes. K-means is a classical clustering algorithm, where the in-cluster sum of squared Euclidean distances of the points (in images, pixel values), w.r.t the cluster centroids, is iteratively minimized. In this work, the K-means is used to define 2 clusters of pixels: negative and positive class. On the other hand, OTSU is a classical thresholding method, that finds a threshold value that minimizes the variance within each of the 2 classes.


\section{Implementation details of the Experimental Evaluation}
\label{sec:experiments}

This section describes the implementation details of the conducted experiments. Firstly, the collected orthomosaics were post-processed to generate representative data of each vineyard. The new dataset is then used in a cross-validation scheme to study the various segmentation networks, as well as the impact of the bands in the segmentation performance.

\subsection{Datasets}
\label{sec:gte}

\begin{figure}[t]
    \centering
    \includegraphics[width=1\columnwidth,trim={0.5cm 0.1cm 0.2cm 0.3cm},clip]{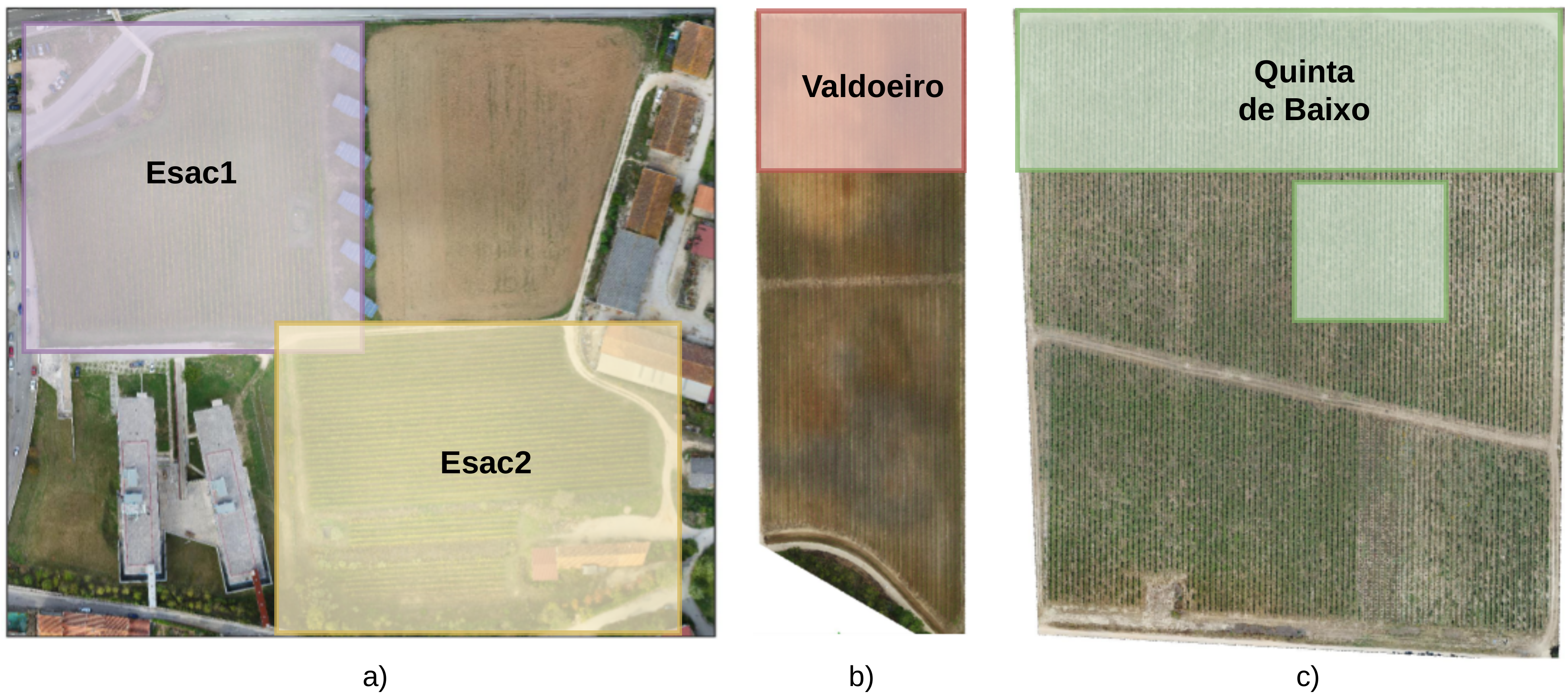}
   \caption{Areas of interest of (a) Coimbra's vineyard plots (ESAC1 and ESAC2) and (b) Valodeiro's plot (Valdoeiro).}
    \label{fig:learning_data}
\end{figure}

The collected orthomosaics from  Esac, Valdoeiro, and Quinta de Baixo were post-processed and a sub-region of each orthomosaic was selected to generate a representative set of each vineyard. The sub-region of each orthomosaic is illustrated in Fig.\,\ref{fig:learning_data}, which represents, in the Valdoeiro and Quinta de Baixo case, the upper region of the orthomosaics. The  ESAC orthomosaic was split into two regions, corresponding to Esac1 and Esac2 (they include distinct vine types).  

For practical reasons, given the limited GPU memory available for training, the orthomosaics of each set were divided into 240$\times$240 sub-images. We note that only the images with at least 1 pixel belonging to the positive class (\ie, corresponding to a vine plant) were used on the training stage.

The resulting dataset comprises thus four sets, denominated by Esac1, Esac2, Valdoeiro and QtaBaixo.  Each set comprises data from the HD-RGB camera and MS sensor, as well as the respective masks. More information about the image distribution among the sets is presented in Table \ref{tab:label_distro}, where P represents the positive class (referring to vine-plants pixels) and N the negative class (referring to non-vineyard pixels).

\subsection{Ground-truth data}

\begin{figure}[t]
    \centering
    \includegraphics[width=1\columnwidth]{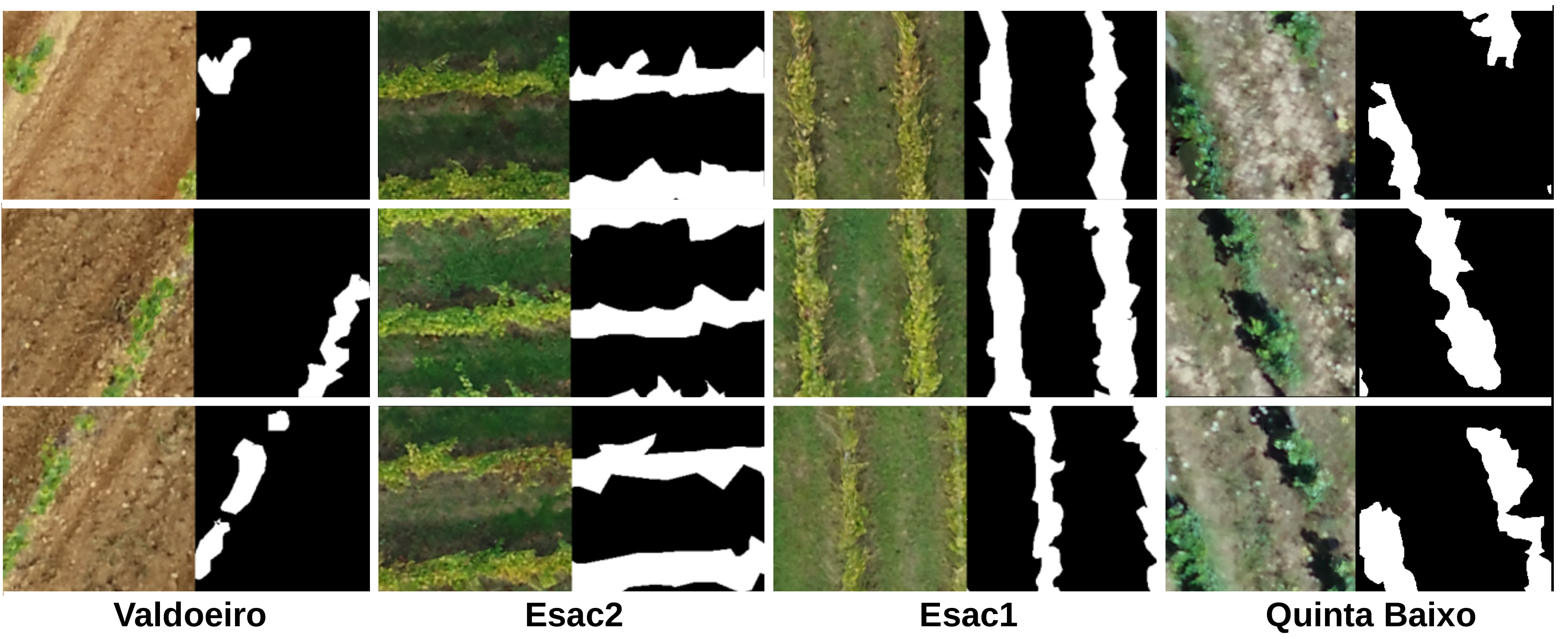}
   \caption{Sub-images and corresponding ground truth masks (240 x 240) used for training and testing.}
    \label{fig:trainingdata}
\end{figure}

In segmentation tasks, the ground truth data correspond to masks. In this work, ground truth masks have been generated in the geospatial space (\ie, orthomosaic and DSM spaces), populating the pixels that belong to vine plants with the positive class (label = 1) and the remaining pixels with a negative class (label = 0) \ie, this is a binary segmentation problem.  The masks were split with the same process as the orthomosaics thus, a sub-mask for each sub-image has been created. Figure \ref{fig:trainingdata} illustrates three sub-image samples of the three areas with their respective sub-masks, and Table \ref{tab:label_distro} contains information regarding image/mask and class distributions of each area of interest.

\begin{table}[t]
  \centering
  \caption{Data and class distributions of each sensor modality where P and N represent respectively, the positive and the negative class fraction available in each set.}
  \label{tab:label_distro}%
  \begin{adjustbox}{max width=\columnwidth}
    \begin{tabular}{l|cc|cc|cc}
    \toprule
          & \multicolumn{2}{c|}{\textbf{I/M}} & \multicolumn{2}{c|}{\textbf{HD}} & \multicolumn{2}{c}{\textbf{MS}} \\
          & \multicolumn{1}{c}{MS} & \multicolumn{1}{c|}{HD} & P     & N     & P     & N     \\
    \midrule
    \midrule
    ESAC1 & 85    & 624   & 0.25  & 0.75      & 0.23  & 0.77  \\
    ESAC2 & 89    & 626   & 0.28  & 0.72      & 0.25  & 0.75   \\
    Valdo. & 150   & 1,196 & 0.07  & 0.93      & 0.08  & 0.92   \\
   QtaBaixo & 120  & 766 & 0.16 & 0.84  & 0.19 &0.81 \\
    \bottomrule
    \end{tabular}%
    \end{adjustbox}
\end{table}%

\subsection{Evaluation and Experiments}

The evaluation procedure adopted in this work was k-fold cross-validation, using the F1-score as performance metric:  

\begin{equation} \label{eq:f1}
    F1  = \frac{2 \, TP}{2\,TP + FP+ FN}
\end{equation}

\noindent where the True Positives (TP) are pixels that were correctly classified as vines; False Positives (FP) are pixels that were wrongly classified as a vine plant; True Negative (TN) are pixels that were correctly classified as background; False Negatives (FN) are pixels that were wrongly classified as background.

In particular, the results from this work were generated based on four non-overlapping subsets defined by: Esac1, Esac2, Valodeiro, and QtaBaixo. Hence, four combinations were generated denoted by T1, T2, T3, and T4, the corresponding data distributions are represented in Table \ref{tab:k-fold}. 

The first three sets (\ie, T1, T2, and T3) are used to conduct the band combination and spatial resolution assessments, while T1, T2, T3 and T4 are used for the comparison of the DL segmentation approaches with classical unsupervised segmentation techniques, as well as for the assessment of the generalization capabilities of these methods.

\subsection{Implementation details and Training}

All experiments were conducted using Python 3.7 and PyTorch, which were set up on a hardware with an NVIDIA GFORCE GTx1070Ti GPU and an AMD Ryzen 5 CPU with 32 GB of RAM.


All networks were initialized, trained, and validated using the same conditions. The networks' weights were initialized using a normal distribution with a mean of 1 and a standard deviation of 0.2. The training was performed using the AdamW optimizer \cite{loshchilov2018decoupled} with a learning rate and a weight decay of 0.000171 and 0.00061, respectively. The loss function was  Pytorch's \textit{BCEWithLogitsLoss}  with the positive class weight set to 5, to compensate the unbalanced class distribution (as can be verified in Table \ref{tab:label_distro}). Data augmentation was also implemented in the form of random rotations with angles between 0 and 180 degrees and random changes in the brightness, contrast, saturation, and hue values. Finally, the networks were trained during 20 epochs, using early stopping to extract the best scores.

\begin{table}[t]
\caption{Image/Mask (I/M) distribution among the training and test set for cross-validation. MS denotes multispectral and HD=high-definition.}
\label{tab:k-fold}
    {\renewcommand{\arraystretch}{1.5}
	\begin{adjustbox}{max width=\columnwidth}
        \begin{tabular}{p{0.04\columnwidth}ccc|ccc}
        \toprule
        & \multicolumn{3}{c|}{\textbf{Training Set}}  & \multicolumn{3}{c}{\textbf{Test Set}} \\ 
        & \textbf{Plots} & \multicolumn{2}{c|}{\textbf{I/M}} & \textbf{Plot} & \multicolumn{2}{c}{\textbf{I/M}}\\  
        & & MS & HD &  & MS & HD\\ 
        \midrule
        \midrule
        T1 & Esac1 \& Esac2          & 174   & 1250  & Valdoeiro & 150 & 1196  \\
        T2 & Esac1 \& Valdoeiro      & 235   & 1820  &  Esac2    & 89  & 626 \\ 
        T3 & Esac2 \& Valdoeiro      & 239   & 1822  &  Esac1    & 85  & 624 \\
        T4  &Esac1 \& Esac2 \& Vald. & 324   & 2446  &  QtaBaixo & 120 & 766     \\
        \bottomrule
        \end{tabular}
    \end{adjustbox}
   }
\end{table}

\section{RESULTS AND DISCUSSION}
\label{sec:results}

\begin{table*}[t]
  \centering
  \caption{Average F1 scores of 5 repetitions. Each repetition was trained with the same parameters: 20 epochs, data augmentation, weight initialization using a normal distribution.}
   \label{tab:results}%
  {\renewcommand{\arraystretch}{1.5}
  \begin{adjustbox}{max width=\textwidth}
    \begin{tabular}{c|cccc|ccccc|ccccc|ccccc}
    \toprule
      \multicolumn{1}{c}{Sensor} & \multicolumn{4}{c}{Bands} & \multicolumn{5}{c}{SegNet} & \multicolumn{5}{c}{U-Net} & \multicolumn{5}{c}{ModSegNet} \\ \hline
    & RGB   & RE & NIR & Th. & T1 & T2 & T3 & Mean & Std & T1 & T2 & T3 & \multicolumn{1}{c}{Mean} &Std & T1 & T2 & T3 & Mean & Std \\
    \midrule
    \midrule
    \multirow{15}{*}{\rotatebox[origin=c]{90}{Multispectral}} 
                    & $\times$  &           &           &               & 0.73  & 0.78  & 0.79  & 0.77  & 0.03  & 0.73  & 0.76  & 0.82  & 0.77  & 0.04  & 0.72  & 0.77  & 0.77  & 0.75  & 0.02 \\
                    &           & $\times$  &           &               & 0.74  & \textbf{0.81} & 0.82  & 0.79  & 0.04  & 0.71  & 0.78  & \textbf{0.85} & 0.78  & 0.06  & 0.65  & \textbf{0.81} & 0.82  & 0.76  & 0.08  \\
                    & $\times$  & $\times$  &           &               & 0.71  & 0.8   & 0.82  & 0.78  & 0.05  & 0.79  & 0.78  & 0.84  & 0.8   & 0.03  & 0.75  & \textbf{0.81} & 0.82  & 0.79  & 0.03  \\
                    &           &           & $\times$  &               & 0.79  & \textbf{0.81} & \textbf{0.83} & \textbf{0.81} & 0.02  & \textbf{0.81} & 0.78  & 0.84  & \textbf{0.81} & 0.02  & 0.74  & \textbf{0.81} & 0.81  & 0.79  & 0.03 \\
                    & $\times$  &           & $\times$  &               & 0.79  & \textbf{0.81} & \textbf{0.83} & \textbf{0.81} & 0.02  & 0.8   & \textbf{0.79} & \textbf{0.85} & \textbf{0.81} & 0.03  & \textbf{0.8} & 0.8   & 0.82  & \textbf{0.81} & 0.01  \\
                    &           & $\times$  & $\times$  &               & \textbf{0.8} & \textbf{0.81} & \textbf{0.83} & \textbf{0.81} & 0.01  & 0.8   & \textbf{0.79} & \textbf{0.85} & \textbf{0.81} & 0.03  & 0.72  & 0.8   & 0.82  & 0.78  & 0.04 \\
                    & $\times$  & $\times$  &$\times$   &               & 0.79  & \textbf{0.81} & \textbf{0.83} & \textbf{0.81} & 0.02  & 0.8   & \textbf{0.79} & \textbf{0.85} & \textbf{0.81} & 0.03  & 0.77  & 0.8   & \textbf{0.83} & 0.8   & 0.02 \\
                    &           &           &           & $\times$      & 0.17  & 0.4   & 0.38  & 0.32  & 0.1   & 0.19  & 0.4   & 0.39  & 0.33  & 0.1   & 0.17  & 0.38  & 0.38  & 0.31  & 0.1 \\
                    & $\times$  &           &           & $\times$      & 0.74  & 0.77  & 0.78  & 0.76  & 0.02  & 0.71  & 0.76  & 0.82  & 0.76  & 0.04  & 0.75  & 0.76  & 0.74  & 0.75  & 0.01 \\
                    &           & $\times$  &           & $\times$      &0.71  & 0.79  & 0.82  & 0.77  & 0.05  & 0.74  & 0.77  & \textbf{0.85} & 0.79  & 0.05  & 0.65  & \textbf{0.81} & 0.81  & 0.76  & 0.08 \\
                    & $\times$  & $\times$  &           &$\times$       & 0.72  & 0.8   & 0.82  & 0.78  & 0.04  & 0.74  & 0.78  & 0.84  & 0.79  & 0.04  & 0.74  & 0.8   & 0.81  & 0.78  & 0.03  \\
                    &           &           & $\times$  & $\times$      & 0.79  & 0.8   & \textbf{0.83} & \textbf{0.81} & 0.02  & 0.78  & \textbf{0.79} & 0.84  & 0.8   & 0.03  & 0.76  & 0.8   & 0.8   & 0.79  & 0.02 \\
                    & $\times$  &           & $\times$  & $\times$      & 0.78  & 0.8   & \textbf{0.83} & 0.8   & 0.02  & 0.79  & \textbf{0.79} & 0.84  & \textbf{0.81} & 0.02  & \textbf{0.8} & 0.79  & 0.81  & 0.8   & 0.01 \\
                    &           & $\times$  &$\times$   & $\times$      & 0.77  & \textbf{0.81} & \textbf{0.83} & 0.8   & 0.02  & 0.79  & \textbf{0.79} & 0.84  & \textbf{0.81} & 0.02  & 0.72  & 0.8   & 0.81  & 0.78  & 0.04 \\
                    & $\times$  & $\times$  & $\times$  & $\times$      & 0.76  & 0.8   & \textbf{0.83} & 0.8   & 0.03  & \textbf{0.81} & \textbf{0.79} & 0.84  & \textbf{0.81} & 0.02  & 0.78  & 0.8   & 0.82  & 0.8   & 0.02 \\
    \midrule
   HD               & $\times$  &           &           &               &  0.73  & 0.85 & 0.85  & 0.81  & 0.06  & 0.75  & 0.82  & 0.91  & 0.83  & 0.07  & 0.75  & 0.83  & 0.89  & 0.82  & 0.06 \\
    \bottomrule
    \end{tabular}%
    \end{adjustbox}
    }
\end{table*}%

This section presents the results and the discussion w.r.t. the objectives of this work. The comparisons are given in terms of the three DL networks, the spatial resolutions, the band combinations, and the classical vs DL-based approaches. The performance of the segmentation approaches are presented and discussed based on quantitative measures as shown in Tables \ref{tab:results} and \ref{tab:comparison}. Additionally, qualitative results are discussed in Section \ref{sec:classical}.

\subsection{Network Comparison and Generalization}\label{sec:generalization}

The results shown in Table\,\ref{tab:results}, which represent the segmentation performance of the DL networks on the cross-validation set T1, T2, and T3 of the various band combinations, suggest that the networks have equivalent overall performance; with SegNet and U-Net having slightly higher and more consistent results over the three subsets.


To achieve more consistent performance, the networks were trained using randomly applied spatial and color augmentation techniques. In particular, empirical evidence showed that augmentation of the brightness, contrast, saturation, and hue values is  essential to achieve higher generalization capability. 

Other transformations such as random rotation and horizontal flipping were also applied but had less effect on the performance. Since vineyards are relatively well structured and have a set of ``natural colors"  characterized by the vine plants, the augmentation techniques were adjusted to match the attributes of vineyards, such as colors and orientations.

The T4 results, given in Table\,\ref{tab:comparison}, were also obtained using augmentations during the training phase, corroborating the usefulness of these techniques for the generalization of the networks. Nevertheless, the results in  Table\,\ref{tab:comparison} also indicate that the networks performed poorly on the NIR band of the T4 set (QtaBaixo test set), despite the augmentation techniques.

\begin{figure}[t]
    \centering
    \includegraphics[width=1\columnwidth]{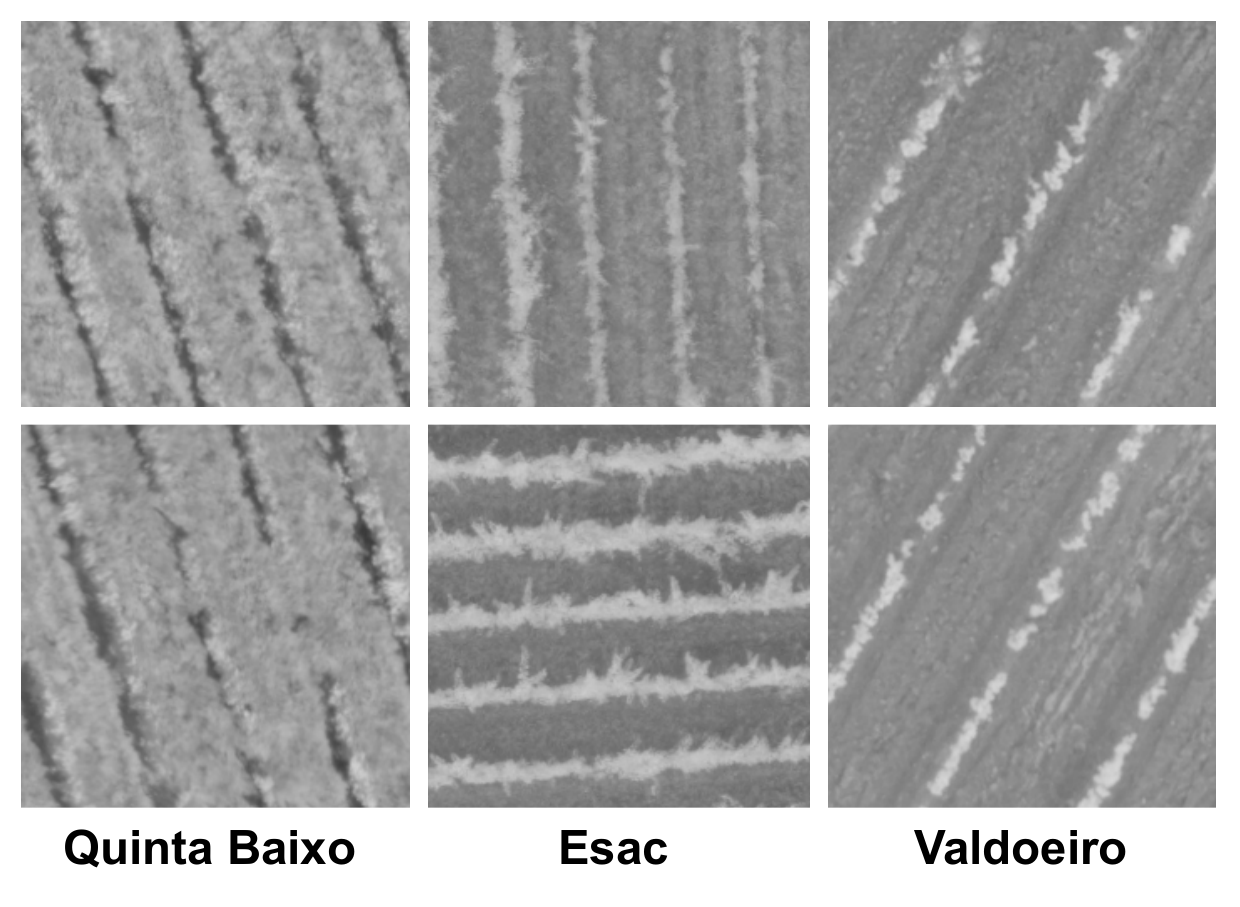}
   \caption{NIR Sub-images of the Quinta de Baixo, Esac and Valdoeiro datasets.}
    \label{fig:nir_bands}
\end{figure}

\begin{figure}[tb]
    \centering
    \includegraphics[width=1\columnwidth]{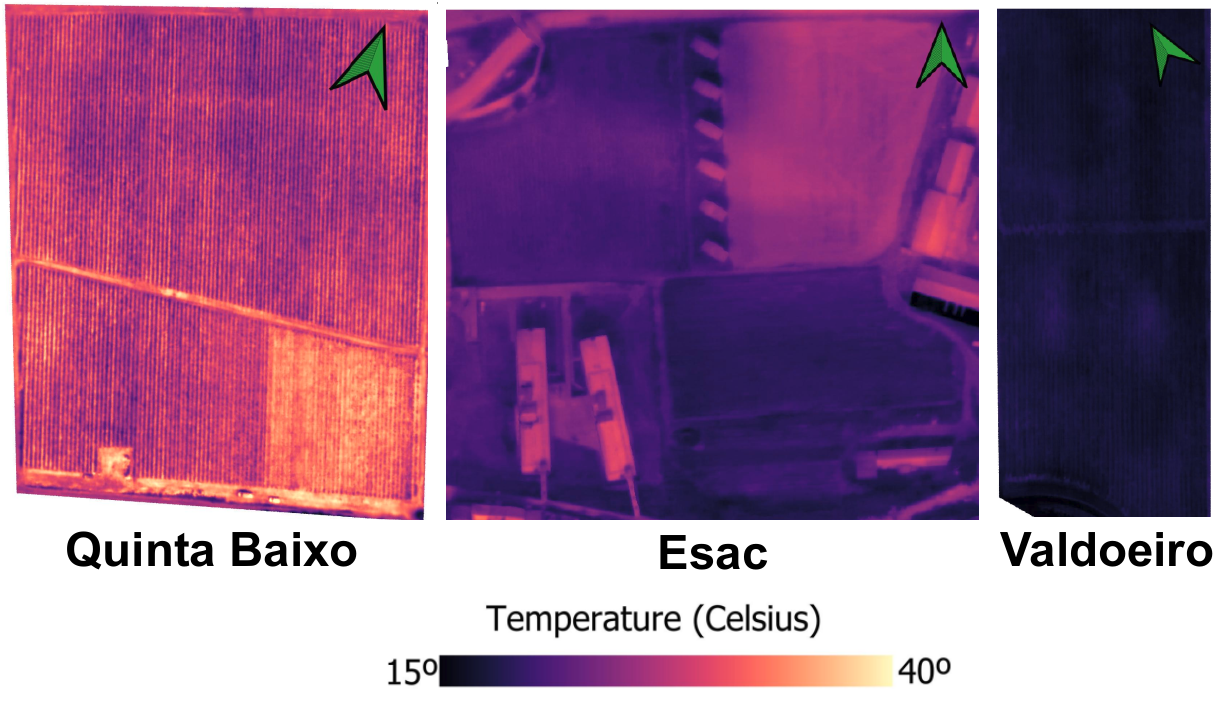}
   \caption{Thermal orthomosaics of Quinta de baixo, Esac, and Valdoeiro sites, where the average temperature at acquisition time was between 28-30\degree C,  around 17\degree C and between 19-20\degree C, respectively.}
    \label{fig:thermal}
\end{figure}

We speculate that a possible cause of the lower performance of the DL networks is grounded in the fact that the datasets were captured under different environmental conditions, namely different environment temperatures. Esac was captured in early autumn with a mean temperature between 19-20\degree C, Valdoeiro was captured in early spring with a mean temperature between 17\degree C, and QtaBaixo was captured in mid-summer with a mean temperature between 28-30\degree C (see Fig.\,\ref{fig:thermal} temperature distribution). In Fig.\,\ref{fig:nir_bands},  NIR images of the three datasets (\ie, Esac, Valdoeiro, and QtaBaixo) are presented, showing clearly that in the Esac and Valdoeiro the vine plant pixels have a higher brightness compared to the remaining pixels. In the QtaBaixo, the vine plant pixels are less highlighted due to a higher overall temperature.  This observation suggests that, despite the NIR band being a valuable information source, it is also highly \CP{suggestible} sensitive to environmental variations such as temperature.  Therefore, we can note that \CP{Meaning that} if not properly handled during the training \ie, \CP{namely} by including more representative data in the training set, using the NIR band may lead to poor results as demonstrated in this study.

\subsection{Image Resolution Comparison}

Table \ref{tab:results} shows, in the first and last row respectively, the F1-scores for different camera resolutions \ie, the RGB bands of the MS and HD cameras. \CP{HD-based F1-scores} We can see that the achieved performance is, in general, higher for the HD camera, which can be partially explained by a larger amount of training data. \CP{available. Note that} As given in Table\,\ref{tab:k-fold}, the HD sets comprise in average 7 times more examples than the MS sets. \CP{: as given in Table\,\ref{tab:k-fold}.}

DL-based approaches are highly data demanding thus, having more data for training with adequate classes distribution, tends to lead to higher performance. However, it is interesting to note that, in some cases like in the T1 cross-validation scenario, different spectral information can be more relevant than extra spatial information. 

\vspace{-0.18cm}

\subsection{ Spectral Band Comparison}

One notable observation from the results is that the NIR spectral band tends to achieve the best results (when compared with modalities of the same resolution), which is in line with the literature. 
Furthermore, using this band alone is sufficient to obtain proficient performance. In some cases, adding other bands to the models do not improve the performance. The thermal band is one of such cases, having very low performance when used alone. A possible reason behind the thermal band having such a poor outcome is due to its low resolution when compared to the other bands as provided in Table \ref{tab:sensors}. 

Despite the high performance of the NIR band, having access to such information requires MS sensors, which are less affordable than their RGB counterparts. Due to the low cost and ease of acquiring, color cameras are very popular among the works in agriculture (as can be seen in Table \ref{sota}). In terms of segmentation performance, when comparing the results of the RGB band with the best-performing band combination, RGB has on average $4.1\%$ lower performance, which can be acceptable when a MS sensor is not an option.
\begin{table}[tb]
  \centering
  \caption{Segmentation performance (F1 scores) of the unsupervised (non-deep) and the deep networks using the NIR, HD-RGB and MS-RGB bands. The results from T1, T2 and T3 of the DL networks were replicated from Table \ref{tab:results} to facilitate the comparison.}
  \label{tab:comparison}
    \begin{tabular}{>{\centering}p{0.03\columnwidth}
                    >{\centering}p{0.14\columnwidth}
                    >{\centering}p{0.08\columnwidth}
                    >{\centering}p{0.08\columnwidth}
                    >{\centering}p{0.08\columnwidth}
                    >{\centering}p{0.1\columnwidth}
                    p{0.1\columnwidth}<{\centering}}
    \toprule
      &  & T1 & T2 & T3 & T4 & Mean \\
    \midrule
    \midrule
    \multirow{5}{*}{\rotatebox[origin=c]{90}{NIR}}
    & OSTSU & 0.52 & 0.76  & 0.67  &  \textbf{0.73} & 0.67   \\
    & KMeans& 0.34 & 0.49  & 0.44  &  0.58     & 0.46   \\ 
    & SegNet      & 0.79          & 0.83              & \textbf{0.81} &  0.63     & \textbf{0.77}  \\
    & U-Net       & \textbf{0.81} & \textbf{0.84}    & 0.78           &  0.66     & \textbf{0.77}      \\
    & ModSeg.   & 0.74          & 0.81              &\textbf{0.81}  &  0.59     & 0.74      \\
    \midrule
    \multirow{5}{*}{\rotatebox[origin=c]{90}{RGB HD}}
    & OSTSU     & 0.55    & 0.63        & 0.55          &  \textbf{0.82}     &   0.64 \\
    & KMeans    & 0.63    & 0.51    & 0.54    &  0.58     &  0.57 \\ 
    & SegNet    & 0.73  & \textbf{0.85}  & 0.85  &  0.76     & 0.80  \\
    & U-Net     & \textbf{0.75}  & 0.82  & \textbf{0.91}  &  0.75     &   \textbf{0.81}   \\
    & ModSeg.   & \textbf{0.75}  & 0.83  & 0.89  &  0.76     &   \textbf{0.81}    \\
    \midrule 
    \multirow{5}{*}{\rotatebox[origin=c]{90}{RGB MS}}
    & OSTSU       & 0.47   & 0.63   & 0.55  &  0.67  &  0.58 \\
    & KMeans      & 0.58      & 0.49   & 0.49  &  0.62 &  0.55 \\ 
    & SegNet      & \textbf{0.73} & \textbf{0.78} & 0.79 &  0.71  & 0.75 \\
    & U-Net       & \textbf{0.73} & 0.76 & \textbf{0.82}  & 0.71  & \textbf{0.76} \\
    & ModSeg.     & 0.72  & 0.77  &0.77 &  \textbf{0.73}   &    0.75   \\
    \bottomrule
    \end{tabular}%
\end{table}%

\begin{figure*}[t]
    \centering
    \includegraphics[width=1\textwidth]{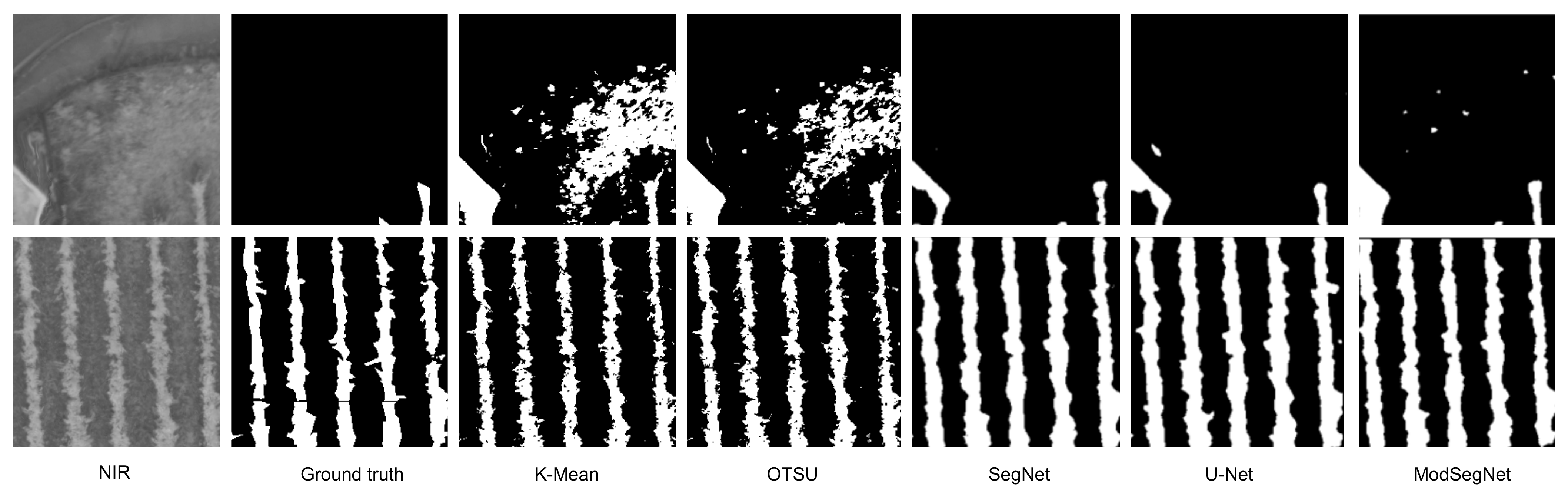}
   \caption{Qualitative prediction masks comparison of DL-based and classical approaches.  The two samples represent: (upper) a corner-case where classical approaches have low performance; and (lower) an ideal case where classical approaches  are  competitive with DL-based approaches.}
    \label{fig:comparison}
\end{figure*}

\subsection{Comparison with Conventional Unsupervised Methods} \label{sec:classical}
The comparison study of the DL networks vs the classical unsupervised segmentation methods is based on the cross-validation sets  T1, T2, T3, and T4, using the NIR and RGB bands (of both HD and MS). The NIR band alone was selected due to leading to the highest performance, while the VIS (RGB) bands, from both HD and MS cameras, were selected due to being widely used in the literature. The classical methods were evaluated under the same conditions as the DL counterparts \ie, by segmenting each sub-image separately instead of the whole orthomosaic.  The results of this comparative study are presented in Table\,\ref{tab:comparison}, where the DL network scores on T1, T2, and T3 sets are replicated from Table \ref{tab:results} to facilitate the comparison. The overall performance of the methods is given by averaging over the 4 subsets.

In general, DL-based methods outperform the classical approaches but, in the NIR band of the T4 set, the DL approaches present a considerably lower performance when compared to the HD-RGB and MS-RGB sets. The potential causes that have led to such performance have been addressed in Section\,\ref{sec:generalization}, being the high environmental temperature during T4's data acquisition probably the main cause. The remaining results show that K-means is completely inadequate for this task, presenting scores near 0.5. OTSU, on the other hand, has a competitive performance in T4, despite struggling in some corner cases where the `positive' class is scarce, contrarily to DL-based approaches, as illustrated in Fig. \ref{fig:comparison}.

Lastly, DL approaches have similar performances on the MS-based sets (\ie MS-RGB and NIR) and achieved better performance, in terms of the average F1-scores, on the HD-RGB bands,  which reinforces the idea that DL-based approaches perform better with more training data.

\section{CONCLUSIONS}
\label{sec.conclusion}

In this work, a new UAV-based MS and HD-RGB dataset was used to train three deep segmentation networks for the task of pixel-wise vineyard segmentation. The aim was to study the responses of the different spectral bands, image resolutions, and segmentation networks when used in this agricultural application. The data was captured from three distinct vineyards at different seasonal stages, all located in the central region of Portugal: Coimbra, Valdoeiro, and Quinta de Baixo.

From the results, three major conclusions can be drown. Firstly, SegNet, U-Net, and Mod-SegNet have equivalent overall performance in vine segmentation. Secondly, the NIR band is essential and generally sufficient to obtain satisfactory performance in most of the datasets. Thirdly, higher image resolution in the HD-RGB spectrum increases the general performance of the DL networks, when compared to the different MS modalities. Lastly, the DL-based networks have in general higher performance than the unsupervised segmentation methods, despite the latter having competitive performances in particular conditions.

The present article makes a good case for the use of this type of dual-camera approach to UAV-based data acquisition, highlighting the clear advantages and disadvantages of each option and discussing, in a thorough and rigorous way, the best semantic segmentation approaches for each scenario. Finally, the  DL-based networks were compared with traditional approaches, underlining the importance of this type of study for real-life precision agriculture applications. For future work, a combination of data acquired from both cameras could be introduced in our analysis of Neural Network performance, as well as some depth information retrieved from the DSMs.

\section*{ACKNOWLEDGMENTS}
This work has been supported by the Portuguese Foundation for Science  and  Technology (FCT) via the projects $AI^{+}Green$ (MIT-EXPL/TDI/0029/2019) and $Agribotics$ (UIDB/00048/2020), and through a PhD grant with the reference  2021.06492.BD. The work of G. Gon\c{c}alves was also supported  by the FCT through the grant UIDB/00308/2020. 
\bibliographystyle{elsarticle-num-names} 
\bibliography{ref}

\end{document}